\documentclass[runningheads]{llncs}

\usepackage[T1]{fontenc}
\usepackage[misc]{ifsym}

\usepackage{mwe}

\usepackage{siunitx}
\usepackage{mathtools}
\usepackage{bbold}
\usepackage{graphicx,caption,subcaption}
\usepackage{kotex}
\usepackage{makecell}
\usepackage{multirow}
\usepackage{amsmath,amssymb,amsfonts}%
\usepackage{IEEEtrantools}
\usepackage{mathrsfs}%
\usepackage[title]{appendix}%
\usepackage{xcolor}%
\usepackage{textcomp}%
\usepackage{manyfoot}%
\usepackage{booktabs}%
\usepackage{algorithm}%
\usepackage{algorithmicx}%
\usepackage{algpseudocode}%
\usepackage{listings}%
\usepackage{url}
\usepackage{hyperref}
\usepackage{wrapfig}
\usepackage{microtype}

\makeatletter
\newcommand{\printfnsymbol}[1]{%
  \textsuperscript{\@fnsymbol{#1}}%
}
\makeatother

\newcommand{\bftab}{\fontseries{b}\selectfont}

\tocauthor{Han-Jun Choi, Hyunsung Kim, Minho Lee, Minchul Jeong, Changjo Kim, Jinsung Yoon, Sang-Ki Ko}
\toctitle{Trajectory Imputation in Multi-Agent Sports with Derivative-Accumulating Self-Ensemble}

\begin{document}

\title{Trajectory Imputation in Multi-Agent Sports with Derivative-Accumulating Self-Ensemble}

\titlerunning{MIDAS}


\author{
Han-Jun Choi\inst{1}\thanks{These authors contributed equally to the paper.}
Hyunsung Kim\inst{2,3}\printfnsymbol{1} \and
Minho Lee\inst{4} \and
Minchul Jeong\inst{5} \and \\
Changjo Kim\inst{3} \and
Jinsung Yoon\inst{3} \and
Sang-Ki Ko\inst{6}\Letter
}
\authorrunning{H. Choi et al.}
%
\institute{
KETI, Seongnam, South Korea\\
\email{hanjun\_c@keti.re.kr} \and
KAIST, Daejeon, South Korea\\
\email{hyunsung.kim@kaist.ac.kr} \and
Fitogether Inc., Seoul, South Korea \\
\email{\{hyunsung.kim,changjo.kim,jinsung.yoon\}@fitogether.com}\and
Saarland University, Saarbrücken, Germany \\
\email{minho.lee@uni-saarland.de} \and
Weflo Inc., Seoul, South Korea \\
\email{mcjeong@weflo.ai} \and
University of Seoul, Seoul, South Korea \\
\email{sangkiko@uos.ac.kr}
}

\maketitle

\begin{abstract}
Multi-agent trajectory data collected from domains such as team sports often suffer from missing values due to various factors. While many imputation methods have been proposed for spatiotemporal data, they are ill-suited for multi-agent sports, where player movements are highly dynamic and interactions evolve over time. To address these challenges, we propose MIDAS (\textbf{M}ulti-agent \textbf{I}mputer with \textbf{D}erivative-\textbf{A}ccumulating \textbf{S}elf-ensemble), a data-efficient framework that imputes multi-agent trajectories with high accuracy and physical plausibility. It jointly predicts positions, velocities, and accelerations via a Set Transformer-based neural network and refines them by recursively accumulating predicted velocity and acceleration values. These predictions are then combined using a learnable weighted ensemble to produce final imputed trajectories. Experiments on three sports datasets show that MIDAS significantly outperforms existing baselines, with particularly large margins in limited-data settings. We also demonstrate its utility in downstream tasks such as estimating total distance and pass success probability. The source code is available at \url{https://github.com/gkswns95/midas.git}.


\end{abstract} 
\section{Introduction}
\label{sec:intro}

\begin{figure}[t!]
\centering
\begin{subfigure}[t]{0.55\textwidth}
    \centering
    \includegraphics[width=\textwidth]{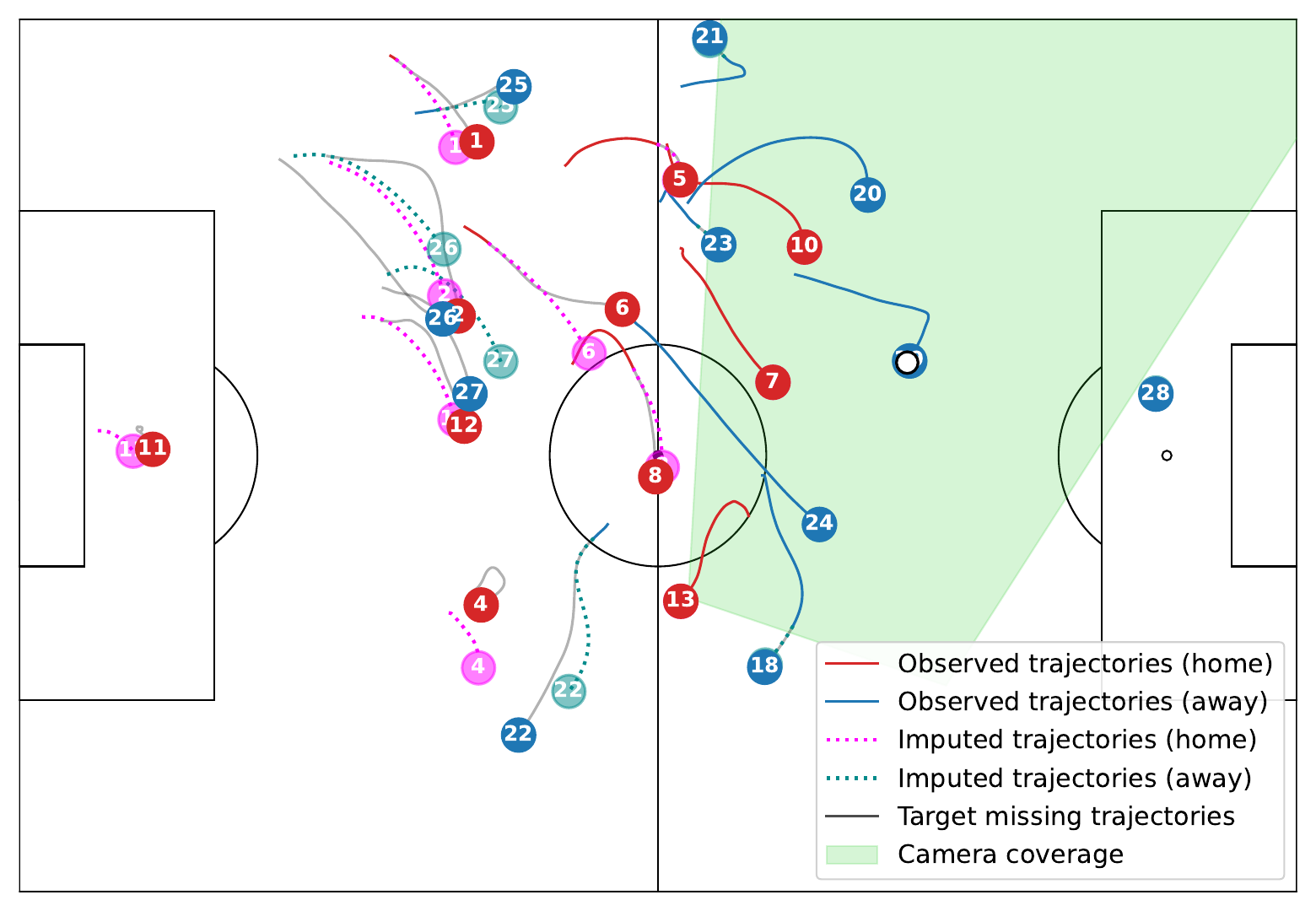}
    \caption{Full-pitch view along with camera coverage}
    \label{fig:first}
\end{subfigure}
\begin{subfigure}[t]{0.34\textwidth}
    \centering
    \includegraphics[width=\textwidth,clip,trim=0 0mm 0mm 0cm]{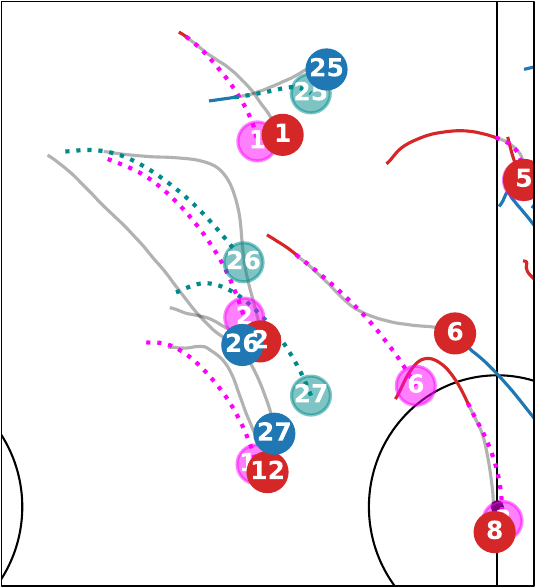}
    \caption{Close-up view}
    \label{fig:second}
\end{subfigure}
        
\caption{Example of imputing unobserved player trajectories from tracking data obtained in a soccer broadcast.}
\label{fig:camera}
\end{figure}

Many spatiotemporal domains, such as transportation, robotics, surveillance, and sports, handle multi-agent trajectory data. While advances in computer vision and sensing technologies have facilitated large-scale trajectory data collection, acquiring complete data remains challenging due to various factors such as signal loss in wearable devices and limitations of the camera's field of view (Fig.~\ref{fig:camera}). This prevalence of missing values calls for the development of effective imputation techniques that can accurately reconstruct missing trajectories.

Though various imputation methods have been proposed for spatiotemporal data, applying them to multi-agent sports remains challenging due to their dynamic nature. In particular, while many of them~\cite{CaoWLZLL18,ChoiL23,LiuYZZY19,MariscaCA22,NieQMMS24,TashiroSSE21} have demonstrated their effectiveness in multi-sensor data, including traffic flow or air quality datasets, they do not account for dynamic interactions between agents. In such fixed-sensor networks, spatial relationships typically remain constant over time. In contrast, multi-agent domains such as team sports involve players with continuously changing positions, requiring a suitable architecture for explicitly modeling these dynamic inter-agent relationships while maintaining permutation-equivariance with respect to the agents.

Furthermore, player motion is governed by biomechanical constraints that impose physical limits on speed, acceleration, and directional change. To accurately model such constraints and generate physically plausible trajectories, imputation models must learn nuanced patterns in movement dynamics. This typically requires a large volume of high-quality training data. However, this poses a significant limitation in the sports domain, where player tracking data is often treated as confidential due to the competitive nature of professional leagues.

Addressing these challenges, this paper proposes MIDAS (\textbf{M}ulti-agent \textbf{I}mputer with \textbf{D}erivative-\textbf{A}ccumulating \textbf{S}elf-ensemble), a framework for multi-agent sports that imputes missing trajectories with high accuracy and data efficiency by explicitly enforcing the physical constraints that real player trajectories should satisfy. First, a neural network equipped with Set Transformers~\cite{LeeLKKCT19} and player-wise bidirectional LSTMs~\cite{HochreiterS97} predicts the positions, velocities, and accelerations of missing players. In addition to this \textit{initial prediction} (IP), it generates alternative estimates through \textit{derivative-accumulating prediction} (DAP), which recursively accumulates the predicted velocity and acceleration values from the nearest observed positions in both forward and backward directions. Finally, the learnable soft voting mechanism combines these three predictions, namely IP and forward/backward DAPs, to produce the final imputed trajectories.

The proposed MIDAS overcomes the aforementioned challenges in two aspects. First, by employing the Set Transformer, it models dynamic inter-agent relationships while ensuring permutation-equivariance with respect to agents. Second, MIDAS jointly predicts positions, velocities, and accelerations, and integrates them through a derivative-accumulating self-ensemble mechanism that explicitly enforces their physical consistency. This design enables the model to capture the underlying patterns of player motion, leading to improved data efficiency.

Experiments across three sports datasets demonstrate that MIDAS consistently outperforms existing baselines in terms of both positional accuracy and physical plausibility of imputed trajectories, with particularly large margins in low-data settings. In addition, we showcase real-world applications of MIDAS, including the approximation of physical (e.g., total distance covered) and contextual (e.g., pass success probability) metrics based on imputed trajectories.

\section{Related Work}
\label{sec:related}

Numerous methods have been proposed for time series and spatiotemporal data imputation. Early models such as BRITS~\cite{CaoWLZLL18}, GRU-D~\cite{ChePCSL18}, and MRNN~\cite{YoonZS19} relied on recurrent models, but suffered from compounding errors due to dependence on their previous predictions. Non-autoregressive frameworks such as NAOMI~\cite{LiuYZZY19}, CSDI~\cite{TashiroSSE21}, and SAITS~\cite{DuCL23} mitigated this issue by enabling parallel imputation across time steps, improving robustness and efficiency. More recently, methods like TIDER~\cite{LiuLCCJ23} and TimesNet~\cite{WuHLZ0L23} were developed to capture temporal patterns such as seasonality or local biases. Other recent approaches further leverage graph structures (GRIN~\cite{CiniMA22}, SPIN~\cite{MariscaCA22}, and NRTSI~\cite{ShanLO23}), information bottlenecks (TimeCIB~\cite{ChoiL23}), or low-rank priors (ImputeFormer~\cite{NieQMMS24}) to enhance imputation performance. Although they perform well on static multi-sensor systems, they are not designed to handle dynamic interactions in multi-agent scenarios.

Meanwhile, several frameworks have proposed dedicated architectures for multi-agent imputation to capture shifting spatial relationships of agents. Notable examples include Graph Imputer~\cite{OmidshafieiHGZ22}, GC-VRNN~\cite{XuBCCF23}, and TranSPORTmer~\cite{CapelleraFRAM24}, which employ dynamic graph or set attention~\cite{LeeLKKCT19} architectures to model time-varying player interactions. More recently, methods tailored for sports data with advanced architectures have been explored, such as U2Diff~\cite{CapelleraRFA25} and Event2Tracking~\cite{HughesHWGFSL25}. However, the former suffers from high computational costs and low data efficiency inherent to diffusion models, while the latter requires manually annotated event data, which are not always available. Interaction modeling has been widely studied in future trajectory forecasting~\cite{GuptaJFSA18,KipfFWWZ18,LiYTC20,SunKZLKWH22,YehSHM19,YuanWOK21}, but these methods are not directly applicable to missing trajectory imputation.
\section{Proposed Framework}

Our study about multi-agent trajectory imputation assumes a scenario where the missing time intervals of players could differ from one another. To elaborate, let the trajectories of $K$ players be $X_{1:T} = \{\mathbf{x}^k_{1:T}\}_{k=1}^K$, where each player $k$'s input features $\mathbf{x}^k_t$ at each time $t$ consist of their $(x,y)$ position $\mathbf{p}^k_t = (p^k_{t,x}, p^k_{t,y})$, velocity $\mathbf{v}^k_t = (v^k_{t,x}, v^k_{t,y})$, and acceleration $\mathbf{a}^k_t = (a^k_{t,x}, a^k_{t,y})$. Here, the velocity and acceleration are calculated from the position values by the following approximations:
\begin{equation}
    \mathbf{v}^k_t \approx \frac{\mathbf{p}^k_t - \mathbf{p}^k_{t-1}}{\Delta{t}}, \quad \mathbf{a}^k_t \approx \frac{\mathbf{v}^k_{t+1} - \mathbf{v}^k_t}{\Delta{t}} 
    \label{eq:vel_accel}
\end{equation}
where $\Delta{t}$ is the difference between adjacent time steps.

In our scenario, each $\mathbf{x}^k_{1:T}$ has missing parts identified by a masking sequence $\mathbf{m}^k_{1:T} = (m^k_1, \ldots, m^k_T)$ where $m^k_t = 1$ if $\mathbf{x}^k_t$ is \emph{observed} and $0$ if it is \emph{missing}.
Then, an imputation model aims to take the incomplete data $\{ \mathbf{m}^k_{1:T} \odot \mathbf{x}^k_{1:T} \}_{k=1}^K$ as input and produce imputed trajectories $\{ \hat{\mathbf{x}}^k_{1:T} \}_{k=1}^K$. Combining these with the observed fragments results in complete trajectories, i.e.,
\begin{equation}
    \tilde{\mathbf{x}}^k_{1:T} = \mathbf{m}^k_{1:T} \odot \mathbf{x}^k_{1:T} + (\mathbb{1}_T - \mathbf{m}^k_{1:T}) \odot \hat{\mathbf{x}}^k_{1:T}, \quad k = 1, \ldots, K. \label{eq:combine}
\end{equation}

\begin{figure}[bth]
\centering
\includegraphics[width=\textwidth, trim={7cm 0cm 2cm 0cm}, clip]{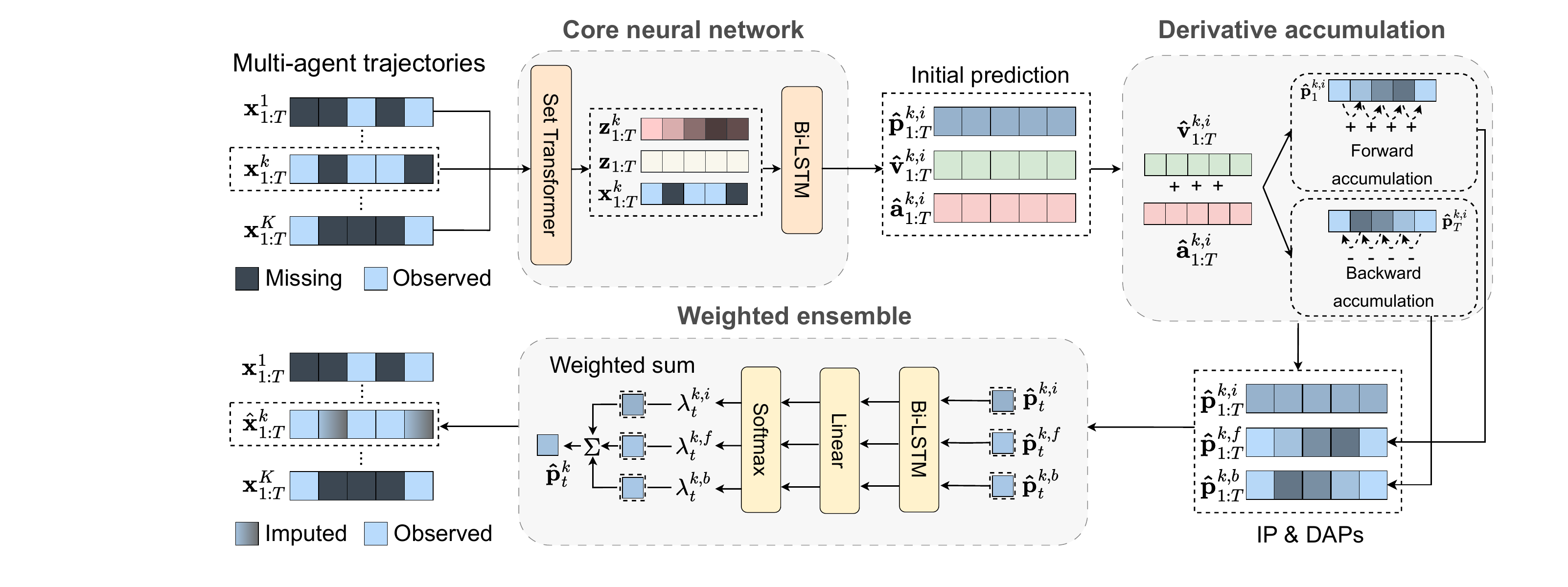}
  \caption{Overview of the proposed framework.}
  \label{fig:overview}
\end{figure}

The novelty of the proposed framework lies in the mechanism of enhancing the model performance by combining positions directly predicted by a neural network and those resulting from accumulating predicted derivatives (i.e., velocity and acceleration values). To elaborate on the details of the proposed mechanism, the remainder of this section consists of the following four parts: Section \ref{se:initial} describes the neural network for initial prediction, Section \ref{se:dap} introduces the derivative accumulation process for alternative predictions, Section \ref{se:hybrid} describes the weighted ensemble mechanism to combine multiple predictions resulting from the previous sections, and Section \ref{se:loss} explains the loss function for model training. See Fig.~\ref{fig:overview} illustrating the overall architecture of our framework.

\subsection{Neural Network-Based Initial Prediction} 
\label{se:initial}
This section describes the neural network architecture that makes \emph{initial prediction} (IP) of imputed trajectories. It takes partially observed trajectories $\{ \mathbf{m}^k_{1:T} \odot \mathbf{x}^k_{1:T} \}_{k=1}^K$ as an input and predicts each player $k$'s full trajectory
\begin{equation}
    \hat{\mathbf{x}}^{k,i}_{1:T} = \{(\hat{p}^{k,i}_{t,x}, \hat{p}^{k,i}_{t,y}, \hat{v}^{k,i}_{t,x}, \hat{v}^{k,i}_{t,y}, \hat{a}^{k,i}_{t,x}, \hat{a}^{k,i}_{t,y})\}_{t=1}^T,
    \label{eq:initial}
\end{equation}
where the superscript $i$ stands for ``initial''.

Since there is generally no inherent order among players in team sports, 
modeling their movements requires ensuring \emph{permutation-equivariance}. That is, permuting the input order should not affect each player’s output, except for applying the same permutation in the output order. Following a recent study on the ball trajectory inference in team sports~\cite{KimCKYK23}, we employ Set Transformer~\cite{LeeLKKCT19} to ensure the permutation-equivariance of outputs.

To be specific, we obtain permutation-equivariant player-wise embeddings $\{ \mathbf{z}^k_t \}_{k=1}^K$ from the encoder of a Set Transformer and a single permutation-invariant embedding $\mathbf{z}_t$ from a full Set Transformer for each time step $t$:
\begin{align}
    (\mathbf{z}_{t}^{1}, \ldots, \mathbf{z}_{t}^{K}) & =  \text{ST-Encoder} (m^1_t \mathbf{x}^1_t, \ldots, m^K_t \mathbf{x}^K_t),  \nonumber \\
    \mathbf{z}_t & =  \text{SetTransformer}
    (m^1_t \mathbf{x}^1_t, \ldots, m^K_t \mathbf{x}^K_t).
    \label{eq:pe_z}
\end{align}
Frame-by-frame application of the Set Transformer to the input features yields embeddings $\{ (\mathbf{z}^1_t, \ldots, \mathbf{z}^K_t, \mathbf{z}_t) \}_{t=1}^T$. Then, bidirectional LSTMs~\cite{HochreiterS97} sharing weights across players extract the sequential information from the concatenated sequence $\{(\mathbf{x}^k_t, \mathbf{z}^k_t, \mathbf{z}_t)\}_{t=1}^T$ per player $k$ by updating joint hidden states:
\begin{equation}
    \mathbf{h}_t^{k,f} = \text{LSTM}^f (\mathbf{x}^k_t, \mathbf{z}^k_t, \mathbf{z}_t ; \mathbf{h}_{t-1}^{k,f}), \quad
    \mathbf{h}_t^{k,b} = \text{LSTM}^b (\mathbf{x}^k_t, \mathbf{z}^k_t, \mathbf{z}_t ; \mathbf{h}_{t+1}^{k,b})
    \label{eq:lstm}
\end{equation}
Lastly, a fully-connected layer decodes the joint hidden state to output a prediction $\hat{\mathbf{x}}^{k,i}_t = \text{FC}(\mathbf{h}_t^{k,f}, \mathbf{h}_t^{k,b})$ at each time $t$. In later sections, we combine it with alternative predictions to get a more accurate final prediction.

\subsection{Derivative-Accumulating Prediction} 
\label{se:dap}
In this section, we start from the fact that players' acceleration values highly vary over time since they are directly related to their stochastic intents. In contrast, velocities are more correlated across neighboring time steps, and positions exhibit even stronger autocorrelation than their derivatives. This implies that accurately predicting acceleration values can lead to more stable and accurate position estimates, provided the physical relationships in Eq.~\eqref{eq:vel_accel} are maintained. However, since the model introduced in Section \ref{se:initial} does not enforce these relationships in its outputs $\{\hat{\mathbf{x}}^{k,i}_{1:T}\}_{k=1}^K$, accurate prediction of the derivatives does not necessarily lead to improved position accuracy.

Taking this into account, we make alternative \emph{derivative-accumulating predictions} (DAP), which enforces the physical relationships in Eq.~\eqref{eq:vel_accel} for improved stability. Specifically, given a missing segment $(t_s, t_e)$ for a player $k$, we recursively predict positions inside the segment by accumulating velocities and accelerations in either direction using the following equation derived from Eq.~\eqref{eq:vel_accel}:
\begin{equation}
    \mathbf{p}_{t+1}^k \approx \mathbf{p}_t^k + \mathbf{v}_{t+1}^k \Delta{t}, \quad
    \mathbf{v}_{t+1}^k \approx \mathbf{v}_t^k + \mathbf{a}_t^k \Delta{t}.
    \label{eq:pva}
\end{equation}
That is, along the forward direction, we start from the observed position $\mathbf{p}^{k}_{t_s}$ by setting $\hat{\mathbf{p}}^{k,f}_{t_s} =  \mathbf{p}^{k}_{t_s}$ and recursively add predicted velocities and accelerations to obtain \emph{forward predictions} $\hat{\mathbf{p}}^{k,f}_t$ for $t \in (t_s, t_e)$ as follows:
\begin{equation}
    \hat{\mathbf{p}}^{k,f}_{t}
    \approx  \hat{\mathbf{p}}^{k,f}_{t-1} + \hat{\mathbf{v}}^{k,i}_{t} \Delta{t}
    \approx  \hat{\mathbf{p}}^{k,f}_{t-1} + (\hat{\mathbf{v}}^{k,i}_{t-1} + \hat{\mathbf{a}}^{k,i}_{t-1} \Delta{t}) \Delta{t}
    \label{eq:pred_forward}
\end{equation}
Likewise, we start from the observed position $\mathbf{p}^{k}_{t_e}$ at the opposite endpoint and recursively subtract the predicted derivatives to obtain {\em backward predictions} $\hat{\mathbf{p}}^{k,b}_t$.

Adopting DAPs instead of initial prediction carries several advantages. First, since the loss between these DAPs and the ground truth penalizes unstable predictions of the velocity and acceleration, minimizing it improves the smoothness of the predicted derivatives. Considering that existing position-oriented imputation models suffer from fluctuating trajectories, these smooth derivatives have a clear advantage in that they result in more plausible positional predictions. Furthermore, enforcing the relationships between the physical quantities imposes an additional inductive bias on the model, making it more data-efficient.

\subsection{Weighted Ensemble of Multiple Predictions}
\label{se:hybrid}

Alhough DAP introduced in Section \ref{se:dap} has clear advantages over IP resulting from Section \ref{se:initial}, it also has a potential drawback known as the \emph{error compounding problem}. Because DAP only relies on the observation at an endpoint as an anchor and the predicted derivatives that are accumulated on the anchor, prediction errors tend to grow as the number of iterations in Eq. \eqref{eq:pred_forward} (or its backward counterpart) increases. In contrast, IP is robust to this problem since it is less sensitive to estimates at certain time steps.

To balance this trade-off, we take a hybrid approach that combines the strengths of both IP and DAP. Rather than exclusively relying on one prediction, it performs a soft voting ensemble by computing a weighted sum of three predictions, IP and forward/backward DAPs. These weights are dynamically learned through an additional player-wise Bi-LSTM, which adapts the contribution of each prediction at each time step.

More specifically, for each player $k$ and time step $t$, we feed the three predictions $\hat{\mathbf{x}}^{k,i}_t, \hat{\mathbf{p}}^{k,f}_t, \hat{\mathbf{p}}^{k,b}_t$ along with the context embeddings $\mathbf{z}^k_t, \mathbf{z}_t$ from Eq. \eqref{eq:pe_z} into a Bi-LSTM that updates its hidden states:
\begin{align}
    \tilde{\mathbf{h}}^{k,f}_t & =  \text{LSTM}^f(\hat{\mathbf{x}}^{k,i}_t, \hat{\mathbf{p}}^{k,f}_t, \hat{\mathbf{p}}^{k,b}_t, \mathbf{z}^k_t, \mathbf{z}_t, \gamma_t; \tilde{\mathbf{h}}^{k,f}_{t-1}), \\
    \tilde{\mathbf{h}}^{k,b}_t & =  \text{LSTM}^b(\hat{\mathbf{x}}^{k,i}_t, \hat{\mathbf{p}}^{k,f}_t, \hat{\mathbf{p}}^{k,b}_t, \mathbf{z}^k_t, \mathbf{z}_t, \gamma_t; \tilde{\mathbf{h}}^{k,b}_{t+1}).
\end{align}
where $\gamma_t = \exp(-\max \{0, \mathbf{W}_{\gamma} \delta_t + \mathbf{b}_{\gamma}\})$ is the temporal decay factor introduced in BRITS \cite{CaoWLZLL18}, indicating the distance of $t$ from observed endpoints. We define $\delta_t = (t-t_s, t_e-t)$ to provide symmetric time gaps for weighting the bidirectional DAPs. Then, a fully-connected layer with a softmax activation returns
\begin{equation}
    (\lambda^{k,i}_t, \lambda^{k,f}_t, \lambda^{k,b}_t) = \text{Softmax}(\text{FC}(\tilde{\mathbf{h}}^{k,f}_t, \tilde{\mathbf{h}}^{k,b}_t))
\end{equation}
that add up to 1. Based on these weights, the model yields a final prediction
\begin{equation}
    \hat{\mathbf{p}}^k_t =
    \lambda^{k,i}_t \hat{\mathbf{p}}^{k,i}_t +
    \lambda^{k,f}_t \hat{\mathbf{p}}^{k,f}_t +
    \lambda^{k,b}_t \hat{\mathbf{p}}^{k,b}_t.
    \label{eq:hybrid_dynamic}
\end{equation}
Combining this final prediction with the observed fragments by Eq. \eqref{eq:combine} results in complete trajectories across the entire period:
\begin{equation}
    \tilde{\mathbf{x}}^k_{1:T} = \mathbf{m}^k_{1:T} \odot \mathbf{x}^k_{1:T} + (\mathbb{1}_T - \mathbf{m}^k_{1:T}) \odot \hat{\mathbf{x}}^k_{1:T},
    \label{eq:combine_final}
\end{equation}
where $\hat{\mathbf{x}}^{k}_t = (\hat{p}^{k}_{t,x}, \hat{p}^{k}_{t,y}, \hat{v}^{k,i}_{t,x}, \hat{v}^{k,i}_{t,y}, \hat{a}^{k,i}_{t,x}, \hat{a}^{k,i}_{t,y})$.

\begin{figure}[b!]
\centering
\includegraphics[width=\textwidth]{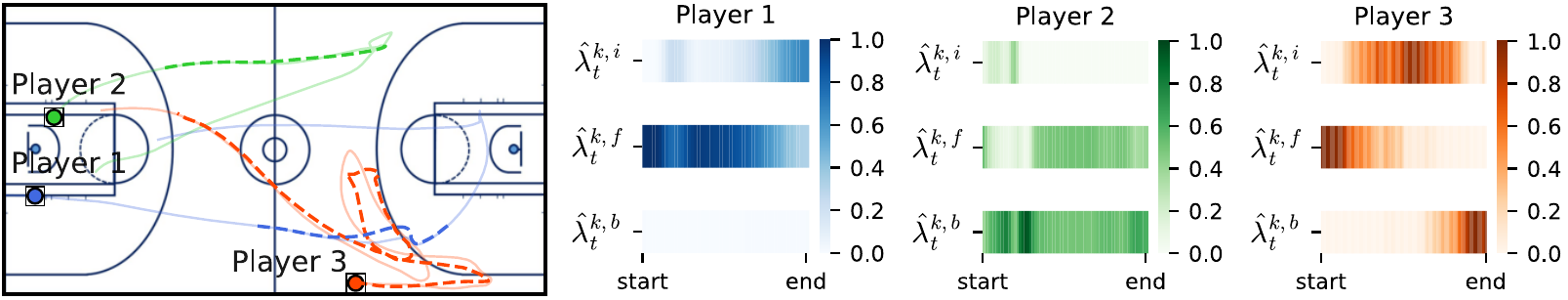}
\caption{Ensemble weight examples for individual imputed trajectories, where circles indicate each player's initial position.}
\label{fig:weights}
\end{figure}

Fig.~\ref{fig:weights} illustrates how MIDAS dynamically adjusts ensemble weights depending on the characteristics of the missing segment. For Player 2, who has a short and stable missing trajectory, the model assigns negligible weight to IP, relying more on DAPs for imputation. In contrast, Player 3's longer and more variable missing trajectory leads to higher IP weights, especially in the middle where DAP errors may accumulate. This highlights MIDAS's ability to improve the final prediction by adaptively combining its components.

\subsection{Loss Function} 
\label{se:loss}
In our framework, improving the accuracy of the prediction at each stage contributes to a more reliable ensemble output. Therefore, we minimize not only the loss between the final ensemble prediction and the true trajectories but also the loss of each auxiliary prediction. Specifically, we compute the mean absolute errors (MAEs) for the IP and DAPs as well as the ensemble prediction, respectively, and train the entire architecture by minimizing the sum of these MAEs. Formally, for the MAE losses $\mathcal{L}^{i}$ of the IP, $\mathcal{L}^{f}$ and $\mathcal{L}^{b}$ for the bidirectional DAPs, and $\mathcal{L}^{h}$ for the ensemble prediction, the model is trained by minimizing
\begin{equation}
    \mathcal{L}^{\rm MIDAS} = \mathcal{L}^{i} + \mathcal{L}^{f} + \mathcal{L}^{b} + \mathcal{L}^{h}.
    \label{eq:total_loss}
\end{equation}
\section{Experiments}
\label{se:experiments}

In this section, we conduct experiments on multiple sports datasets to evaluate the performance of MIDAS and its generalizability across different sports.

\subsection{Data Preparation}
\label{se:data}
In the experiments, we independently trained and evaluated models on three public datasets collected from popular team sports: soccer, basketball, and American football. The soccer dataset is provided by Metrica Sports\footnote{\url{https://github.com/metrica-sports/sample-data}} and contains tracking data for 22 players collected across three matches. For basketball, we use the first 100 matches of SportsVU NBA dataset\footnote{\url{https://github.com/linouk23/NBA-Player-Movements}}, containing trajectories of 10 players per match. The American football dataset is from the Kaggle competition\footnote{\url{https://www.kaggle.com/competitions/nfl-big-data-bowl-2021}} and is based on NFL's Next Gen Stats. We adopt its preprocessed version\footnote{\url{https://github.com/lupalab/NRTSI/tree/main/codes_stochastic}} provided by the NRTSI paper~\cite{ShanLO23}, which contains 9,543 five-second time series of six offensive players. The original sampling rates of the three datasets are 25Hz, 25Hz, and 10Hz, respectively, but we downsample all datasets to 10Hz for consistency. As model inputs, we use 200 frames (20 seconds) per sequence for soccer and basketball and 50 frames (5 seconds) for American football.

\subsection{Missing Scenarios}
\label{se:scenarios}
To evaluate the model performance on various missing patterns, we consider the following three scenarios that may occur during data acquisition processes:
\begin{enumerate}
    \item Uniform missing: All players have missing values at the same time interval. Note that among the baselines, NAOMI~\cite{LiuYZZY19} and NRTSI~\cite{ShanLO23} are designed to only handle this scenario and are not capable of the following other scenarios.
    \item Agent-wise missing: Individual players have different missing intervals.
    \item Broadcasting camera: A virtual camera follows the ball and only captures the players inside the camera view, resulting in missing values for the remaining players as shown in Fig.~\ref{fig:camera}. Following Graph Imputer~\cite{OmidshafieiHGZ22}, we conduct experiments only on the soccer dataset for this scenario.
\end{enumerate}

\begin{figure}[t!]
    \centering
    \includegraphics[width=0.6\textwidth]{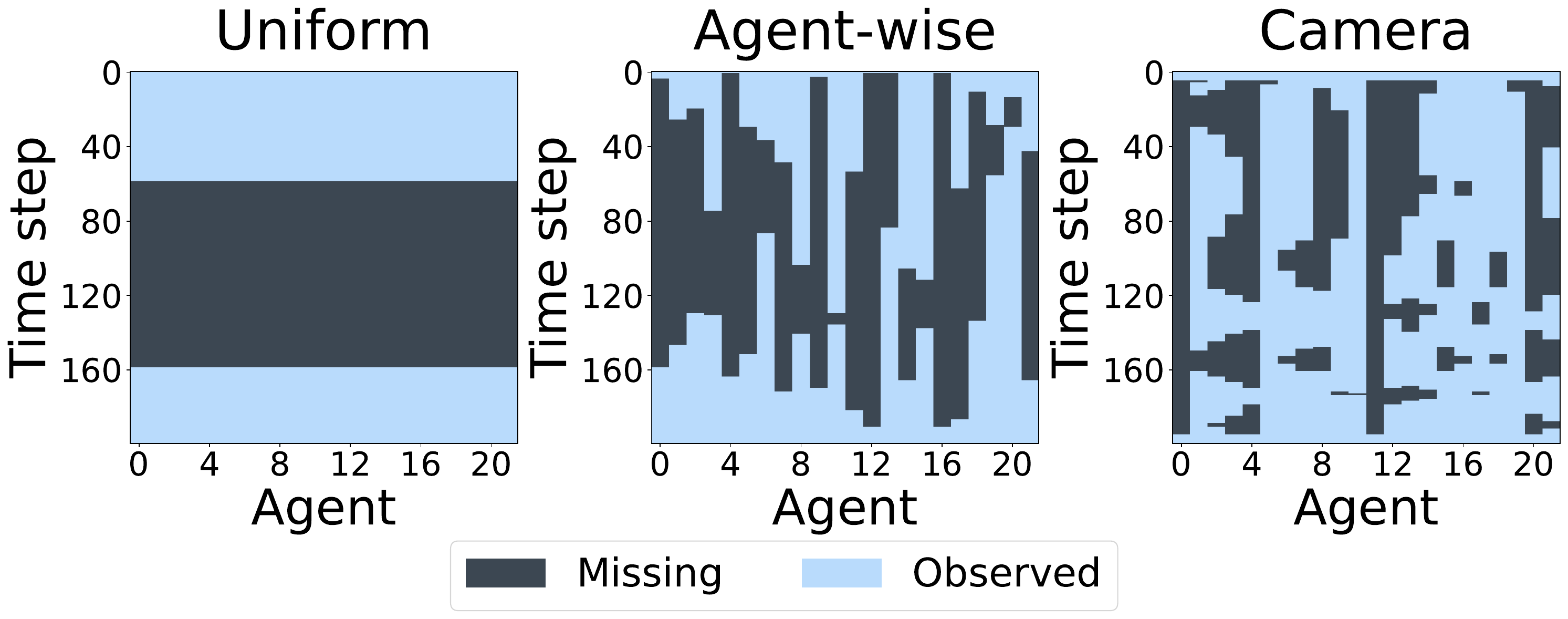}
    \caption{Masking matrix examples for three different missing scenarios.}
    \label{fig:mask}
\end{figure}

Fig.~\ref{fig:mask} illustrates examples of masking matrices from the soccer dataset used in each missing scenario. Since our task is to impute trajectories given observed data before and after the missing intervals, we ensure that the first and last five frames of each sequence are always observable, following previous work~\cite{LiuYZZY19,OmidshafieiHGZ22,ShanLO23}. During training, we apply a dynamic missing rate ranging from 0.1 to 0.9, while testing is conducted with a fixed missing rate of 0.5 to encourage the model to generalize across various missing rate scenarios. Additional evaluations with different missing rates are presented as the ablation study in Section~\ref{se:ablation}.

\begin{table}[t!]
    \centering
    \caption{Details on the three sports datasets.}
    \begin{tabular}{l|cr|cr|cr}
        \toprule
        & \multicolumn{2}{c|}{{\bf Soccer}} & \multicolumn{2}{c|}{{\bf Basketball}} & \multicolumn{2}{c}{{\bf A. Football}} \\
        {\bf Split} & {\bf Matches} & {\bf Frames} & {\bf Matches} & {\bf Frames} & {\bf Matches} & {\bf Frames} \\
        \midrule
        Training    & 2   & 65,014 & 70 & 1,621,835  & --- & 425,000 \\
        Validation  & 0.5 & 20,104 & 10 & 216,118    & --- & 52,150  \\
        Test        & 0.5 & 21,242 & 20 & 468,885    & --- & --- \\
        \bottomrule
    \end{tabular}
    \label{tab:data}
\end{table}

\subsection{Baseline Models and Evaluation Metrics}
In the experiments, we compare the imputation performance of MIDAS with several baselines, including naive methods such as linear interpolation (LI) and cubic spline (CS), as well as deep learning models such as BRITS~\cite{CaoWLZLL18}, NAOMI~\cite{LiuYZZY19}, NRTSI~\cite{ShanLO23}, CSDI~\cite{TashiroSSE21}, Graph Imputer (GI)~\cite{OmidshafieiHGZ22}, and ImputeFormer (IF)~\cite{NieQMMS24}. To evaluate the impact of our derivative-accumulating self-ensemble mechanism, we also implement a baseline that only uses the initial prediction (IP) described in Section \ref{se:initial} and is trained with the loss $\mathcal{L}^i$ from Section \ref{se:loss}. For models such as BRITS, NAOMI, and CSDI that do not preserve permutation-equivariance of input players, we sort trajectories by the sum of their average $x$ and $y$ coordinates to ensure permutation robustness. We compare these baselines using two evaluation metrics: (1) \emph{position error} (PE) indicating the average Euclidean distance between the true and predicted positions and (2) \emph{step change error} (SCE)~\cite{LiuYZZY19,ShanLO23} defined as the average absolute difference between the variance of the true and predicted velocities to assess the trajectories' physical plausibility.

\subsection{Main Experimental Results}
The resulting Table~\ref{tab:results} shows that the proposed MIDAS consistently outperforms other baselines in both positional accuracy (PE) and physical plausibility (SCE). Linear interpolation (LI) and cubic spline (CS) offer simple yet competitive results in some scenarios. Notably, deep learning baselines such as BRITS, NAOMI, and CSDI, which are not designed for multi-agent domains, often fail to exceed these naive baselines. Similarly, Graph Imputer (GI) performs worse than LI and CS in many cases, as previously reported in its original article~\cite{OmidshafieiHGZ22}.

By contrast, our model for initial prediction (IP) already surpasses most baselines, demonstrating its strength in modeling dynamic multi-agent interactions. Furthermore, comparing IP and MIDAS highlights the effectiveness of our self-ensemble mechanism. By effectively leveraging the complementary strengths of initial predictions and alternative derivative-accumulating predictions, it leads to superior performance even compared to state-of-the-art methods such as ImputeFormer (IF) in most scenarios.

\begin{table}[t!]
    \centering
    \caption{Performance of imputation methods on different datasets and scenarios.}
    \resizebox{1.\columnwidth}{!}{%
    \begin{tabular}{lccccccccccc}\toprule
    & & \multicolumn{10}{c}{{\bf Method}}\\ 
        {\bf Scenario} & {\bf Metric} & LI & CS & BRITS & NAOMI & NRTSI & CSDI & GI & IF & IP & MIDAS  \\ \midrule
        \multicolumn{12}{l}{{\bf Soccer}} \\
        \hspace{0.2cm}Uniform & PE & 3.8406 & 2.2085 & 7.4859 & 4.5343 & 3.1791 & 3.4295 & 4.6511 & 2.0898 & 1.4563 & \bftab{1.3205} \\
        & SCE & 0.1299 & 0.0867 & 3.9089 & 3.9793 & 0.0854 & 0.1586 & 0.1191 & 0.0815 & 0.1488 & \bftab{0.0516} \\
        \hspace{0.2cm}Agent-wise & PE& 5.0752 & 11.4647 & 5.7266 & --- & --- & 4.0279 & 5.6011 & 2.5798 & 2.0755 & \bftab{1.9832} \\
        & SCE & 0.1631 & 0.2939 & 2.9627 & --- & --- & 0.1305 & 0.1508 & 0.0976 & 0.1057 & \bftab{0.0535} \\
        \hspace{0.2cm}Camera & PE& 3.1083 & 1.9209 & 7.4208 & --- & --- & 3.5181 & 3.6512 & 2.2151 & 1.4879 & \bftab{1.2296} \\
        & SCE & 0.0993 & 0.0547 & 4.1967 & --- & --- & 0.2132 & 0.0934 & 0.3149 & 0.1554 & \bftab{0.0374} \\\midrule\midrule
        \multicolumn{12}{l}{{\bf Basketball}}\\
        \hspace{0.2cm}Uniform & PE & 3.3481 & 2.3114 & 2.9085 & 1.5254 & 2.5291 & 2.2558 & 2.8305 & 1.3622 & 0.9801 & \bftab{0.9727} \\ 
        & SCE & 0.1483 & 0.1025 & 1.0521 & 0.3230 & 0.0734 & 0.0631 &  0.1066 & 0.0531 & \bftab{0.0432} & 0.0438 \\
        \hspace{0.2cm}Agent-wise & PE & 4.4992 & 10.3857 & 2.4238 & --- & --- & 2.3471 & 2.5859 & \bftab{1.3345} & 1.3832 & 1.3862 \\ 
        & SCE & 0.1787 & 0.2715 & 0.5397 & --- & --- &  0.0563 & 0.0700 & 0.0485 & \bftab{0.0373} & 0.0381 \\\midrule\midrule
        \multicolumn{12}{l}{{\bf American Football}}\\
        \hspace{0.2cm}Uniform & PE & 0.8897 & 0.7448 & 1.7990 & 0.9692 & 0.5158 & 0.5558 & 0.8899 & 0.3673 & 0.2073 & \bftab{0.1542} \\
        & SCE & 1.1063 & 0.9463 & 10.9459 & 2.3112 & 0.2989 & 0.4905 & 1.1023 & 0.2858 & 0.1990 & \bftab{0.1126} \\
        \hspace{0.2cm}Agent-wise & PE & 1.5128 & 1.2041 & 1.7527 & --- & --- & 0.6182 & 1.5128 & 0.3944 & 0.2383 & \bftab{0.2104} \\
        & SCE & 1.0641 & 0.5306 & 10.6807 & --- & --- & 0.4288 & 1.0631 & 0.1869 &  0.1180 & \bftab{0.0967} \\\bottomrule
    \end{tabular}
    }
    \label{tab:results}
\end{table}

\begin{table}[t!]
\centering
\caption{Performance of the top-3 methods on the basketball dataset when trained with full data (70 games) versus limited data (3 games, 71,428 frames).}
\resizebox{.7\columnwidth}{!}{%
\begin{tabular}{lc|ccc|ccc}
\toprule
& & \multicolumn{3}{c|}{\bf Limited Data} & \multicolumn{3}{c}{\bf Full Data} \\
{\bf Scenario} & {\bf Metric} & IF & IP & MIDAS & IF & IP & MIDAS \\
\midrule
Uniform     & PE  & 1.6741 & 1.1868 & \bftab{1.1438} & 1.3622 & 0.9801 & \bftab{0.9727}\\
            & SCE & 0.0628 & 0.1483 & \bftab{0.0493} & 0.0531 & \bftab{0.0432} & 0.0438 \\
Agent-wise  & PE  & 1.8876 & 1.6414 & \bftab{1.5994} & \bftab{1.3345} & 1.3832 & 1.3862 \\
            & SCE & 0.0645 & 0.0452 & \bftab{0.0439} & 0.0485 & \bftab{0.0373} & 0.0381 \\
\bottomrule
\end{tabular}
}
\label{tab:limited_data}
\end{table}

\begin{figure}[t!]
\centering
\begin{subfigure}[b]{0.32\textwidth}
 \centering
 \includegraphics[width=\textwidth]{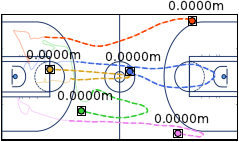}
 \caption{Ground Truth}
 \label{fig:traj_target}
\end{subfigure}
\begin{subfigure}[b]{0.32\textwidth}
 \centering
 \includegraphics[width=\textwidth]{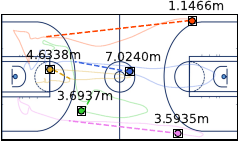}
 \caption{Linear Interpolation}
 \label{fig:traj_linear}
\end{subfigure}
\begin{subfigure}[b]{0.32\textwidth}
 \centering
 \includegraphics[width=\textwidth]{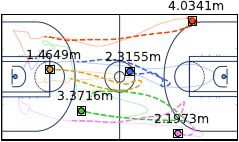}
 \caption{NRTSI}
 \label{fig:traj_nrtsi}
\end{subfigure}
\begin{subfigure}[b]{0.32\textwidth}
 \centering
 \includegraphics[width=\textwidth]{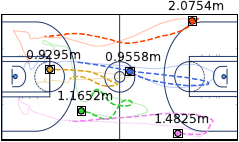}
 \caption{CSDI}
 \label{fig:traj_csdi}
\end{subfigure}
\begin{subfigure}[b]{0.32\textwidth}
 \centering
 \includegraphics[width=\textwidth]{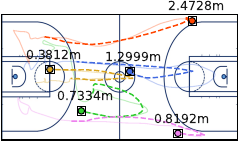}
 \caption{ImputeFormer}
 \label{fig:traj_imputeformer}
\end{subfigure}
\begin{subfigure}[b]{0.32\textwidth}
 \centering
  \includegraphics[width=\textwidth]{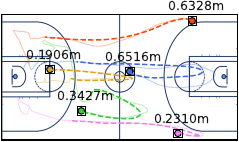}
 \caption{MIDAS (ours)}
 \label{fig:traj_midas}
\end{subfigure}
\caption{Ground truth (solid) and imputed trajectories (dashed) for a uniform missing scenario in basketball. Circles indicate the players' initial positions, and the annotated values represent average position errors over the missing interval.}
\label{fig:traj}
\end{figure}

Such observations are also evident in Fig.~\ref{fig:traj}, where LI, NRTSI, and CSDI often generate unrealistic trajectories, either overly linear or changing direction too frequently, resulting in erratic and implausible motion patterns. In contrast, IF and MIDAS produce trajectories that closely resemble true player movements, as indicated by the relatively low PEs over the missing interval.

Meanwhile, an interesting observation emerges in the basketball dataset, where IP and IF show performance comparable to or slightly better than MIDAS in some cases. We attribute this to the larger amount of training data available (70 games), which may reduce the relative advantage of MIDAS. In contrast, for soccer and American football, where training data are more limited, MIDAS’s derivative-based self-ensemble mechanism demonstrates its data efficiency by significantly outperforming other methods.

To investigate whether this data efficiency holds for the basketball dataset, we measure model performance on an additional setting where models are trained using only 3 games (comparable in scale to the training data used for the soccer dataset). As shown in Table~\ref{tab:limited_data}, all models experience a performance drop when trained on limited data. However, the degradation is much less severe for MIDAS, which eventually outperforms both IF and IP under this setting. This observation underscores the practical utility of MIDAS in real-world sports analytics, where obtaining large amounts of complete tracking data is often difficult due to the competitive and commercial nature of professional sports.

\subsection{Ablation Studies}
\label{se:ablation}
In addition to the main experiments, we conducted two ablation studies to further investigate the model's behavior across varying difficulties and the impact of its features and components on the performance.

\paragraph{Model Behavior Across Varying Difficulties.}
To better understand how MIDAS responds to varying levels of difficulty, we conduct an ablation study analyzing the model’s behavior across different missing lengths. For the agent-wise missing scenario in the soccer and basketball datasets, we divide trajectory segments in the test set into three groups based on missing length.

\begin{table}[b!]
\centering
\caption{Position errors and ensemble weights of each MIDAS component in Eq.~\eqref{eq:hybrid_dynamic} across different missing lengths within 200-frame (\SI{20}{s}) trajectory segments. Segments are grouped into the short, medium, and long thirds based on missing length, where the mean and standard deviation of the number of missing frames in each group are reported in the table.}
\resizebox{1.\columnwidth}{!}{
\begin{tabular}{lccrrrr}
\toprule
\textbf{Sports} & \textbf{Category} & \textbf{Missing Frames} &
\multicolumn{1}{c}{$\hat{\mathbf{p}}^{k,i}_t$ ($\hat{\lambda}^{k,i}_t$)} &
\multicolumn{1}{c}{$\hat{\mathbf{p}}^{k,f}_t$ ($\hat{\lambda}^{k,f}_t$)} &
\multicolumn{1}{c}{$\hat{\mathbf{p}}^{k,b}_t$ ($\hat{\lambda}^{k,b}_t$)} &
\multicolumn{1}{c}{$\hat{\mathbf{p}}^{k}_t$} \\
\midrule
\multirow{3}{*}{Soccer}
    & Short   & \phantom{0}$33.30 \pm 15.72$  & 0.1379 (0.0001) & 0.0742 (0.6501) & 0.0783 (0.3498) & \bftab{0.0504} \\
    & Medium  & \phantom{0}$90.42 \pm 19.17$  & 0.7939 (0.0003) & 0.7492 (0.6110) & 0.7624 (0.3887) & \bftab{0.7004} \\
    & Long    & $173.41 \pm 21.70$ & 2.7359 (0.0002) & 2.7017 (0.5034) & 2.7186 (0.4964) & \bftab{2.6082} \\
\midrule
\multirow{3}{*}{Basketball}
    & Short   & \phantom{0}$33.86 \pm 16.11$  & 0.0444 (0.1596) & 0.0424 (0.5216) & 0.0426 (0.3187) & \bftab{0.0376} \\
    & Medium  & \phantom{0}$90.65 \pm 19.02$  & 0.5395 (0.1783) & 0.5483 (0.4809) & 0.5482 (0.3406) & \bftab{0.5312} \\
    & Long    & $172.60 \pm 21.20$ & 1.8104 (0.1801) & 1.8288 (0.4313) & 1.8258 (0.3985) & \bftab{1.8005} \\
\bottomrule
\end{tabular}
}
\label{tab:ablation_misslen}
\end{table}

\begin{table}[b!]
\centering
\caption{Position errors of each MIDAS component in Eq. \eqref{eq:hybrid_dynamic}, when trained using different subsets of features.}
\resizebox{.7\columnwidth}{!}{
\begin{tabular}{llrrrr}
\toprule
\textbf{IP Features} & \textbf{DAP Features} &
\multicolumn{1}{c}{$\hat{\mathbf{p}}^{k,i}_t$} &
\multicolumn{1}{c}{$\hat{\mathbf{p}}^{k,f}_t$} &
\multicolumn{1}{c}{$\hat{\mathbf{p}}^{k,b}_t$} &
\multicolumn{1}{c}{$\hat{\mathbf{p}}^{k}_t$} \\
\midrule
Position        & Velocity          & 4.0008 & 2.8203 & 2.8261 & 2.5549 \\
Position        & Vel. \& accel.    & 2.9119 & 2.6973 & 2.6858 & 2.4954 \\
Pos. \& vel.    & Velocity          & 1.6600 & 1.5742 & 1.5590 & 1.4025 \\
Pos. \& vel.    & Vel. \& accel.    & 1.6235 & 1.5731 & 1.5670 & 1.4013 \\
All features    & Velocity          & 1.6713 & 1.4985 & 1.5595 & 1.2644 \\
All features    & Vel. \& accel.    & \bftab{1.4963} & \bftab{1.4122} & \bftab{1.3982} & \bftab{1.2296} \\
\bottomrule
\end{tabular}
}
\label{tab:ablation_deriv}
\end{table}

As shown in Table~\ref{tab:ablation_misslen}, the position error increases with longer missing segments, and the advantage of DAP over IP diminishes. When the missing length is short, the DAP components outperform the initial prediction (IP) by a wide margin, resulting in substantial performance gains through the ensemble. However, as the missing length increases, the uncertainty of player motion makes DAP less reliable, reducing its advantage. This trend is reflected in the ensemble weights as well: $\hat{\lambda}^{k,i}_t$ increases and $\hat{\lambda}^{k,f}_t$ decreases as the missing length grows.

In addition, the advantage of DAP over IP varies by sport. In soccer, DAP maintains its superiority over IP even for long segments, and the IP component receives almost zero weight in the ensemble, indicating that MIDAS relies entirely on DAP. In contrast, in basketball, IP surpasses DAP for medium and long segments. This is reflected in noticeably higher $\hat{\lambda}^{k,i}_t$, suggesting its greater influence in ensemble prediction. This difference likely stems from the nature of basketball, where the variability of player motion is higher relative to the small court size, making long-range accumulation of derivatives more error-prone.

\paragraph{Impact of Derivative Features.}
To empirically justify the use of derivative features in IP and DAP, we compared the performance of MIDAS on the broadcasting camera scenario in soccer with those trained without accelerations (i.e., using positions and velocities) and even without velocities (i.e., only using positions) of observed trajectories. Furthermore, we configured IP to predict only positions and velocities, and DAP to estimate positions only based on predicted velocities. We then compared these ``velocity-only'' DAP models with their counterparts that also incorporate predicted accelerations as in Eq.~\eqref{eq:pred_forward}.

Table~\ref{tab:ablation_deriv} shows that providing velocities and accelerations as input features for IP clearly improves the imputation performance. We attribute this to the neural network architecture for IP, where Set Transformers independently encode multi-agent contexts at each time step via Set Transformers, and player-wise Bi-LSTMs link temporal information. Providing derivatives as input allows the Set Transformers to utilize information of adjoining time steps, resulting in more comprehensive context embeddings compared to using positions alone.

In addition, we observe consistent performance improvements when predicted accelerations are incorporated into DAP alongside velocities (i.e., rows 2, 4, and 6 versus 1, 3, and 5, respectively). We attribute this to the fact that including accelerations enables DAP to capture the finer dynamics of player motion, mitigating the drift caused by error compounding in velocity-only DAP.

\subsection{Time Complexity Analysis}
While achieving high imputation accuracy is important, real-world deployment requires efficient processing without excessive computational overhead. To assess the feasibility of real-time use, we analyze both the theoretical and empirical time complexity of the proposed MIDAS framework.

\begin{table}[b!]
\centering
\caption{Empirical analysis of time complexity on soccer and basketball data. Each row corresponds to an imputation method, with the number of parameters, average inference time per 20-second window, and total processing time for half a soccer match and a full basketball match in the test set described in Table~\ref{tab:data}.}
\begin{tabular}{l|rrr|rrr}
\toprule
& \multicolumn{3}{c|}{\bf Soccer} & \multicolumn{3}{c}{\bf Basketball} \\
{\bf Method} & \multicolumn{1}{c}{\bf Params} & \multicolumn{1}{c}{\bf Inference} & \multicolumn{1}{c|}{\bf Total} & \multicolumn{1}{c}{\bf Params} & \multicolumn{1}{c}{\bf Inference} & \multicolumn{1}{c}{\bf Total} \\
\midrule
NRTSI        & 84,090,968 & \SI{893.07}{ms} & \SI{84.40}{s}
             & 84,017,192 & \SI{906.74}{ms} & \SI{100.55}{s} \\
CSDI         & 414,209    & \SI{394.44}{ms} & \SI{39.68}{s} 
             & 413,825    & \SI{324.70}{ms} & \SI{37.68}{s} \\
ImputeFormer & 1,290,978  & \SI{6.19}{ms}   & \SI{4.92}{s}
             & 1,060,578  & \SI{5.68}{ms}   & \SI{3.85}{s} \\
MIDAS (ours) & 3,945,579  & \SI{24.32}{ms}  & \SI{8.22}{s}
             & 3,945,579  & \SI{21.58}{ms}  & \SI{5.71}{s} \\
\bottomrule
\end{tabular}
\label{tab:complexity}
\end{table}

From a theoretical standpoint, we consider three key parameters: the number of time steps $T$, the hidden dimension $m$, and the number of players $K$. It takes $O(TK^2m)$ for Set Transformer used in the initial prediction, $O(TKm^2)$ for Bi-LSTMs employed in the initial prediction and weighted ensemble. In total, $O(TK^2m + TKm^2) \approx O(TKm^2)$ as $m \gg K$. Since Bi-LSTM adopts a recurrent structure, the entire computation process proceeds over $O(T)$ steps, proportional to the length of the sequence.

In addition to this analysis, we empirically evaluated the inference time of MIDAS and several baseline models using an NVIDIA TITAN RTX GPU with 24GB memory. As summarized in Table~\ref{tab:complexity}, although MIDAS is slightly slower than ImputeFormer~\cite{NieQMMS24}, it can process an entire 40-minute game (half-match for soccer and full game for basketball) within 10 seconds. This demonstrates that MIDAS operates well within real-time constraints, enabling its potential use in practical scenarios such as live match analysis or broadcasting augmentation.
\section{Applications}
\label{se:use_cases}
In Section \ref{se:experiments}, we evaluate model performance primarily based on metrics related to missing periods in sports data. In practice, however, what is more important in this domain is the quality of the trajectories for the entire period, as they can be utilized in diverse domain-specific downstream tasks. As examples, we present two promising applications in the soccer domain: approximating physical statistics and pass success probability from incomplete tracking data.

\subsection{Approximation of Physical Statistics for Load Management}
\label{se:match_stats}

\begin{table}[b!]
\centering
\caption{Players' physical statistics estimated by different models and their mean absolute percentage errors (MAPE) for the soccer test data.}
\begin{tabular}{l|rr|rr}
\toprule
& \multicolumn{2}{c|}{\textbf{Distance (m)}} & \multicolumn{2}{c}{\textbf{Sprints}} \\
\textbf{Method} & \multicolumn{1}{c}{\textbf{Mean}} & \textbf{MAPE} & \textbf{Mean} & \textbf{MAPE} \\
\midrule
Ground Truth    & 11,093.5  & \multicolumn{1}{c|}{---} & 41.49 & \multicolumn{1}{c}{---} \\
\midrule
Linear Interp.  & 10,167.8  & 8.46\% & 38.89 & 6.32\%   \\
Cubic Spline    & 10,686.3  & 3.73\% & 38.85 & 6.73\%   \\
BRITS           & 10,979.2  & 2.76\% & 59.89 & 53.62\%  \\
CSDI            & 11,343.0  & 2.77\% & 44.20 & 14.71\%  \\
Graph Imputer   & 8,972.1 & 19.15\% & 37.85 & 9.80\% \\
ImputeFormer    & 11,441.7 & 3.22\% & 50.25 & 26.29\%  \\
MIDAS (ours) & {\bftab 10,922.4} & \bftab{1.58}\% & {\bftab 40.71} & \bftab{4.95}\%\\
\bottomrule
\end{tabular}
\label{tab:stats}
\end{table}

In this section, we explore how accurately our method can estimate statistics for a given period when imputed trajectories are combined with known observations. Specifically, we compare the \emph{total distance} covered by a player and the \emph{number of sprints} estimated by each method, as they are widely used as indicators for players' physical performance or fitness. We first compute velocities from the observed/imputed positions based on Eq. \ref{eq:vel_accel} and obtain speed values by calculating the norms of these velocity vectors. To make the best estimation from given positional predictions, we remove outliers whose speed is larger \SI{12}{\meter\per\second} or whose norm of the acceleration exceeds \SI{8}{\meter\per\square\second} and replace the values by linear interpolation. Also, we smoothen the resulting speed signal by applying a Savitzky-Golay filter~\cite{SavitzkyG64}. After preprocessing speed signals, we compute the distance covered by each player by summing the speed values multiplied by $\Delta{t} = 0.1$s. For the latter, if a player runs faster than \SI{6}{\meter\per\second} for consecutive frames, we detect his/her movement during the frames as \emph{sprint} and count the number of such sprints the player made during the given period.

For evaluation, we use the soccer test data consisting of the half of a match and assume the broadcasting camera scenario. Since players played for different time periods, we normalize each player's statistics by 90 minutes and calculate the averages of such normalized values estimated by MIDAS and other baselines, respectively. Note that players who ran fewer than two sprints during the half were excluded from every evaluation.

According to Table \ref{tab:stats}, MIDAS provides accurate estimates close to the ground truth. Especially considering that almost all baselines either suffer from inaccurate distance measures (linear interpolation, cubic spline, and Graph Imputer) or overestimate speed spikes (BRITS, CSDI, and ImputeFormer), it is obvious that our model takes clear advantage of smooth prediction of velocities. In a nutshell, our framework is practical in that it can provide reliable statistics with incomplete tracking data, which originally require complete player trajectories.

\begin{figure}[b!]
\centering
\begin{subfigure}[b]{0.40\textwidth}
 \centering
 \includegraphics[width=\textwidth,height=.60\textwidth]{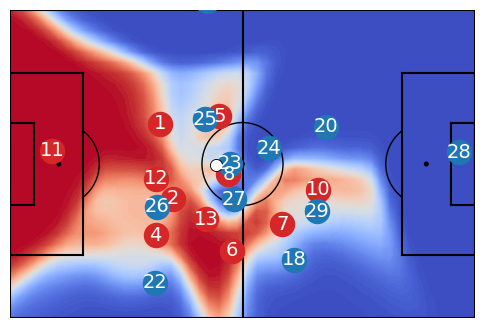}
 \caption{Ground Truth}
 \label{fig:pc_target}
\end{subfigure}
\begin{subfigure}[b]{0.40\textwidth}
 \centering
 \includegraphics[width=\textwidth,height=.60\textwidth]{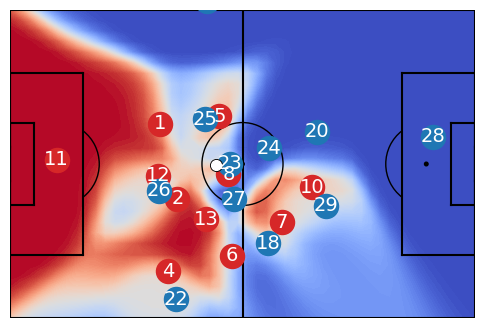}
 \caption{CSDI}
 \label{fig:pc_csdi}
\end{subfigure}
\begin{subfigure}[b]{0.40\textwidth}
 \centering
 \includegraphics[width=\textwidth,height=.60\textwidth]{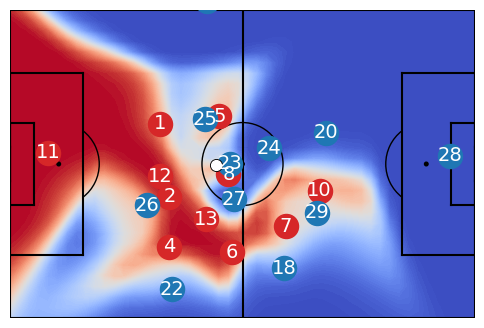}
 \caption{ImputeFormer}
 \label{fig:pc_imputeformer}
\end{subfigure}
\begin{subfigure}[b]{0.40\textwidth}
 \centering
 \includegraphics[width=\textwidth,height=.60\textwidth]{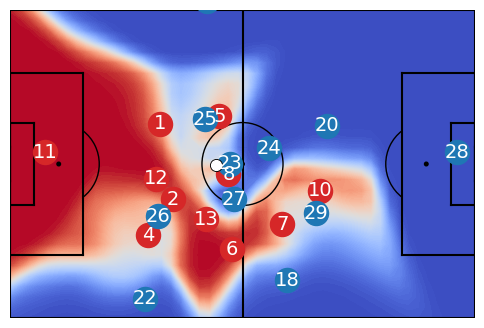}
 \caption{MIDAS (ours)}
 \label{fig:pc_midas}
\end{subfigure}
\caption{Pass success probability maps based on true and imputed player positions resulting from different methods in a partially-observable camera setting. Darker red and blue regions indicate a higher probability of the left and right teams gaining possession, respectively, if the ball is passed to those areas.}
\label{fig:pitch_control}
\end{figure}

\subsection{Approximation of Pass Success Probability for Spatial Analysis}
\label{se:pitch_control}
One representative example of leveraging player tracking data for match analysis is Pitch Control~\cite{SpearmanBDHP17}, which estimates the probability that a pass to each location on the pitch would be successful. Such pass success probabilities for different destinations are typically visualized as a heat map overlayed on the pitch, so that domain experts can evaluate players' positioning and decision-making for both actual and hypothetical passes to different locations on the pitch.

Thus, to demonstrate the applicability of imputation models in actual match analysis, we compare the Pitch Control maps generated using the player positions imputed by different methods. In the situation in Fig.~\ref{fig:pitch_control} as an example, player 23 of the blue team has the ball, while his teammates 22 and 26 are making forward runs towards the open spaces on the left flank and in the center behind the red defensive line, respectively. The ground truth map (Fig.~\ref{fig:pc_target}) indicates that the blue team has a slightly higher control probability in both areas, reflecting reasonable pass success opportunities to these players, who are onside. However, inaccurate imputations by CSDI (Fig.~\ref{fig:pc_csdi}) and ImputeFormer (Fig.~\ref{fig:pc_imputeformer}) lead to issues such as players being mispredicted in offside positions or pass success probabilities being over- or underestimated in critical areas. In contrast, MIDAS (Fig.~\ref{fig:pc_midas}) produces predictions that closely follow the actual player dynamics, resulting in probability maps that more accurately reflect the true game situation. This example instantiates how our framework facilitates more reliable downstream analysis by providing accurate imputation results.

\section{Conclusions}
This paper proposes MIDAS, a framework for imputing missing values in multi-agent trajectories with high accuracy and physical plausibility. MIDAS combines a permutation-equivariant neural network for initial trajectory prediction with a self-ensemble mechanism that incorporates derivative-based alternative predictions to refine imputation results and enforce physical consistency. Experiments on three team sports datasets under various missing scenarios demonstrate the effectiveness of our approach in generating trajectories with higher positional accuracy and improved reality than existing baselines. While this study focuses on team sports, we believe the proposed framework is applicable to other spatiotemporal domains. Future work will explore extending MIDAS to additional domains such as autonomous driving and crowd simulation, as well as its application to downstream tasks that require complete and reliable trajectory data.
\section*{Acknowledgments}
Han-Jun Choi was supported by Institute of Information \& Communications Technology Planning \& Evaluation (IITP) grant funded by the Korea government (MSIT) (No. RS-2022-II220608) and Sang-Ki Ko was supported by the National Research Foundation of Korea (NRF) grant funded by MSIT (No. RS-2023-00208094).

\bibliographystyle{splncs04}
\bibliography{ref}

\end{document}